\documentclass[conference]{IEEEtran}
\IEEEoverridecommandlockouts
\usepackage{cite}
\usepackage{amsmath,amssymb,amsfonts}
\usepackage{algorithm}
\usepackage{algorithmic}
\usepackage[font=footnotesize]{caption}
\usepackage{graphicx}
\usepackage{textcomp}
\usepackage{xcolor}
\usepackage{enumitem}
\usepackage{dblfloatfix}
\usepackage{titlesec}
\usepackage[square,numbers]{natbib}
\usepackage{url}

\usepackage{tikz}
\newcommand*\circled[1]{\tikz[baseline=(char.base)]{
            \node[shape=circle,draw, color=black,inner sep=0.8pt, minimum size=2pt] (char) {#1};}}
            
\newcommand{\rpoint}[1]{\circled{{\fontfamily{pcr}\selectfont\footnotesize{#1}}}}

\titlespacing\section{0pt}{0.5\baselineskip}{0.3\baselineskip}
\titlespacing\subsection{0pt}{0.4\baselineskip}{0.2\baselineskip}
\titlespacing\subsubsection{0pt}{0.2\baselineskip}{0.1\baselineskip}

\def\BibTeX{{\rm B\kern-.05em{\sc i\kern-.025em b}\kern-.08em
    T\kern-.1667em\lower.7ex\hbox{E}\kern-.125emX}}

\usepackage{fancyhdr,lipsum}
\fancypagestyle{firstpage}{
  \fancyhf{}
  \fancyhead[C]{To appear at the 2021 International Joint Conference on Neural Networks (IJCNN), Virtual Event, July 2021.}
  \fancyfoot[C]{\thepage}
}

\begin{document}

\title{DVS-Attacks: Adversarial Attacks on Dynamic Vision Sensors for Spiking Neural Networks\\
\vspace*{-10pt}}


\author{\IEEEauthorblockN{Alberto Marchisio\textsuperscript{1,*}\thanks{*These authors contributed equally to this work.}, Giacomo Pira\textsuperscript{2,*}, Maurizio Martina\textsuperscript{2}, Guido Masera\textsuperscript{2}, Muhammad Shafique\textsuperscript{3}}
\IEEEauthorblockA{\textit{\textsuperscript{1}Technische Universit{\"a}t Wien, Vienna, Austria}\ \ \ \textit{\textsuperscript{2}Politecnico di Torino, Turin, Italy}\ \ \ \textit{\textsuperscript{3}New York University, Abu Dhabi, UAE}} 
\IEEEauthorblockA{\textit{Email: alberto.marchisio@tuwien.ac.at, giacomo.pira@studenti.polito.it}}
\IEEEauthorblockA{\textit{\{maurizio.martina, guido.masera\}@polito.it, muhammad.shafique@nyu.edu}}\\
\vspace*{-40pt}}
\maketitle
\thispagestyle{firstpage}





\begin{abstract}

Spiking Neural Networks (SNNs), despite being energy-efficient when implemented on neuromorphic hardware and coupled with event-based Dynamic Vision Sensors (DVS), are vulnerable to security threats, such as adversarial attacks, i.e., small perturbations added to the input for inducing a misclassification. 
Toward this, we propose \textit{DVS-Attacks}, a set of stealthy yet efficient adversarial attack methodologies targeted to perturb the event sequences that compose the input of the SNNs. First, we show that noise filters for DVS can be used as defense mechanisms against adversarial attacks. Afterwards, we implement several attacks and test them in the presence of two types of noise filters for DVS cameras. The experimental results show that the filters can only partially defend the SNNs against our proposed \textit{DVS-Attacks}. Using the best settings for the noise filters, our proposed \textit{Mask Filter-Aware Dash Attack} reduces the accuracy by more than 20\% on the DVS-Gesture dataset and by more than 65\% on the MNIST dataset, compared to the original clean frames. The source code of all the proposed \textit{DVS-Attacks} and noise filters is released at \url{https://github.com/albertomarchisio/DVS-Attacks}.
\end{abstract}

\begin{IEEEkeywords}
Spiking Neural Networks, SNNs, Deep Learning, Adversarial Attacks, Security, Robustness, Defense, Filter, Perturbation, Noise, Dynamic Vision Sensors, DVS, Neuromorphic, Event-Based.\\
\end{IEEEkeywords}

\vspace*{-15pt}

\section{Introduction}

Spiking Neural Networks (SNNs) represent energy-efficient learning models in a wide variety of machine learning applications, e.g., autonomous driving~\cite{Zhou2020SNNAD}, healthcare~\cite{Gonzalez2020SNNhealthcare}, and robotics~\cite{Tang2018SNNrobotics}. Unlike traditional (i.e., non-spiking) Deep Neural Networks (DNNs), the SNNs are more closely related to the human brain's processing~\cite{Kasinski2011IntroSNN}. Indeed, the event-based communication between neurons makes them biologically plausible. Moreover, SNNs are appealing for being implemented in resource-constrained embedded systems~\cite{Capra2020SurveyDNN}, due to a good combination of power/energy efficiency and real-time classification performance. In fact, compared to the equivalent DNN implementations, SNNs exhibit a lower computational load, as well as a reduction in the latency, by leveraging the spike-based communication between neurons~\cite{Deng2020ComparisonANNSNN}.

Efficient SNNs are typically implemented on a specialized neuromorphic hardware~\cite{Schuman2017SurveyNeuromorphic}, which is able to exploit the asynchronous communication mechanism between neurons and the event-based propagation of the information through layers. These characteristics led to an increasing interest in developing neuromorphic architectures like IBM TrueNorth~\cite{Merolla2014Truenorth} and Intel Loihi~\cite{Davies2018Loihi}. Another advancement in the field of neuromorphic hardware has come from the new generation of event-based camera sensors, such as the Dynamic Vision Sensor (DVS)~\cite{Lichtsteiner2006DVS}. Unlike a classical frame-based camera, the DVS emulates the behavior of the human retina, by recording the information in form of a sequence of spikes, which are generated every time a change of light intensity is detected. 
The event-based behavior of these sensors pairs well with SNNs implemented onto the neuromorphic hardware, i.e., the output of a DVS camera can be used as the direct input of an SNN to process events in real-time.

\subsection{Target Research Problem and Scientific Challenges}

Different security threats challenge the correct functionality of DNNs and SNNs. The DNN trustworthiness has been extensively investigated in recent years~\cite{Shafique2020RobustMLDnT}, highlighting that one of the most critical issues is the adversarial attacks, i.e., small and imperceptible input perturbations to trigger misclassification~\cite{Szegedy2014IntriguingNN}. Although some initial studies have been conducted~\cite{Bagheri2018AdvTrainingSNN}\cite{Marchisio2019SNNAttack}\cite{Sharmin2019ACA}\cite{Liang2020ExploringAA}, the SNN trustworthiness is a relatively new and unexplored problem. More specifically, DVS-based systems have not yet been investigated for SNN security. Moreover, the methods for defending SNNs against such adversarial attacks can be inspired from the recent advancements of the defense mechanisms for DNNs, where studies have focused on adversarial learning algorithms~\cite{Madry2017TowardsDL}, loss/regularization functions~\cite{Zhang2019TradeoffRobustnessAccuracy}, and image preprocessing~\cite{Khalid2019FAdeML}. 
The latter approach basically consists of suppressing the adversarial perturbation through dedicated filtering. Noteworthy, for the SNN-based systems feeded by DVS signals, the attacks and preprocessing-based defense techniques for frame-based sensors cannot be directly applied due to differences in the signal properties. Therefore, specialized noise filters for DVS sensors~\cite{LinaresBarranco2019FilterDVS} must be employed.

\begin{figure}[h]
    \centering
    \includegraphics[width=\linewidth]{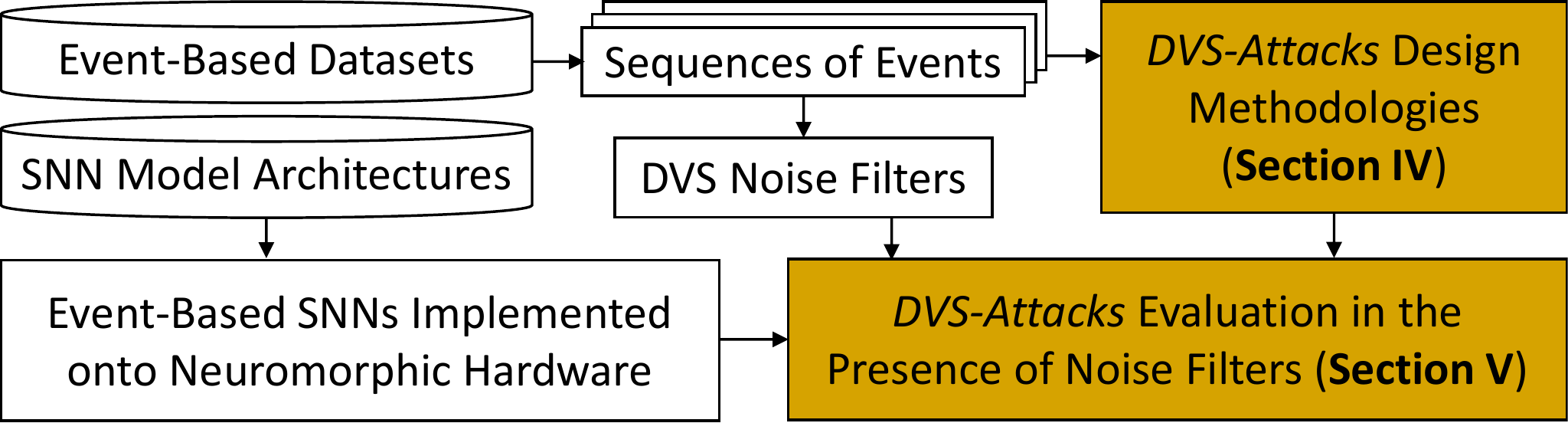}
    \caption{Overview of the steps involved in this work. Our novel contributions are highlighted in colored boxes.}
    \label{fig:novel_contributions}
\end{figure}

As per our knowledge, the generation of adversarial attacks for DVS signals is an unexplored and open research problem. Towards this, we propose \textit{DVS-Attacks}, a set of adversarial attack methodologies for DVS signals, and test them in scenarios where noise filters are employed as a defense mechanism against them. Since the DVS cameras contain also the temporal information, the generation of adversarial perturbation is technically different w.r.t. traditional adversarial attacks on images, where only the spatial information is considered. Hence, the temporal information needs to be leveraged for developing the attack and defense mechanisms. The steps involved in this work are visualized in Fig.~\ref{fig:novel_contributions}.

\subsection{Motivational Case Study}

\begin{figure}[h]
    \centering
    \includegraphics[width=\linewidth]{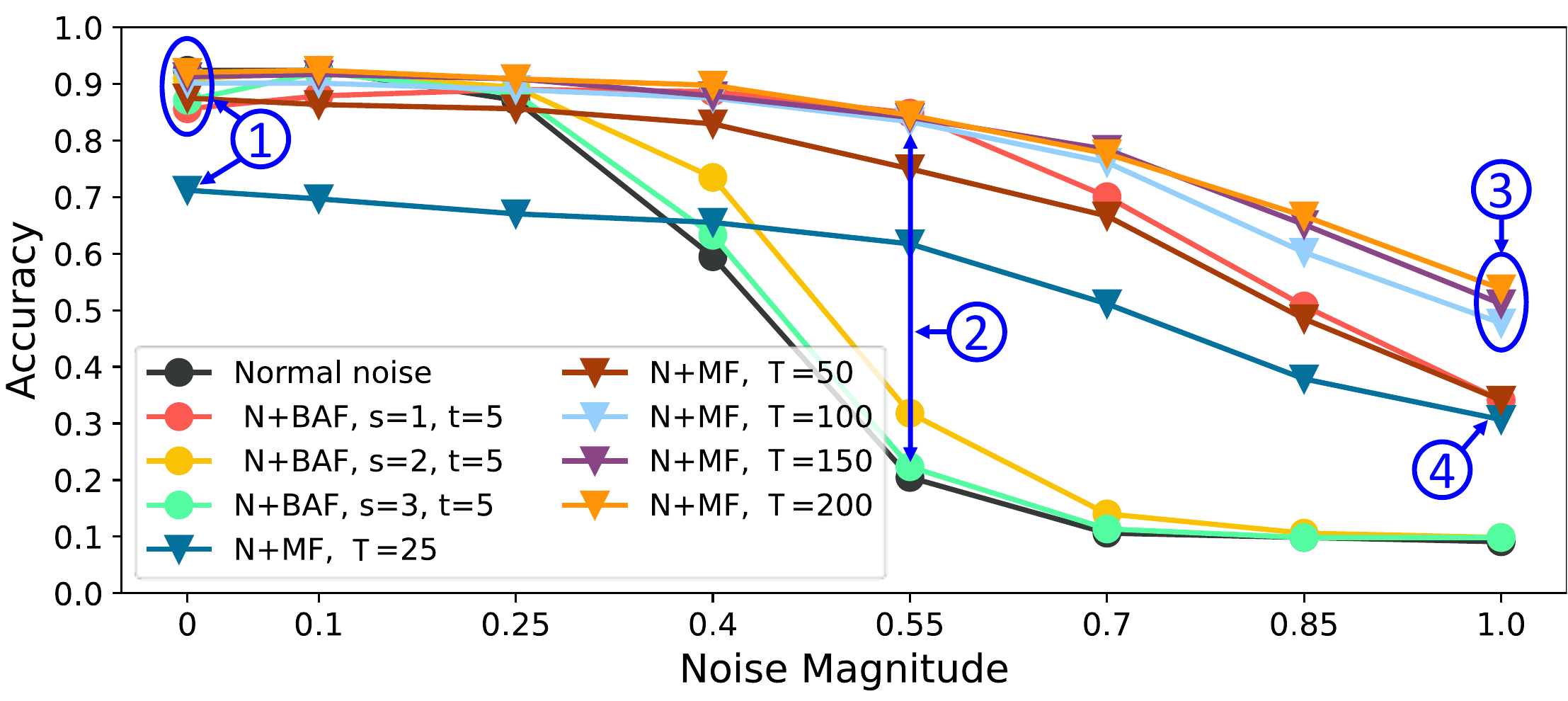}
    \vspace*{-10pt}
    \caption{Analyzing the impact of applying the normally-distributed noise to the DVS-Gesture dataset, in the presence of \textit{BAF} and \textit{MF} noise filters.}
    \label{fig:noise}
\end{figure}

As a preliminary study for motivating our research in the above-discussed directions, we perform the following experiments. We trained a 4-layer SNN with 2 convolutional layers and 2 fully-connected layers, for the DVS-Gesture dataset~\cite{Amir2017DVSgesture} using the SLAYER method~\cite{Shrestha2018SLAYER} in a DL-workstation equipped with two Nvidia GeForce RTX 2080 Ti GPUs. For each frame of events, we perturb the testing dataset by injecting normally-distributed random noise and measure the classification accuracy. Moreover, to mitigate the effect of the perturbations, the \textit{Background Activity Filter} (\textit{BAF}) and the \textit{Mask Filter} (\textit{MF})
of~\cite{LinaresBarranco2019FilterDVS} are applied, with various filter parameters. The accuracy results w.r.t. different noise magnitude are shown in Fig.~\ref{fig:noise}. As indicated by pointer~\rpoint{1} in Fig.~\ref{fig:noise}, the filter may potentially reduce the accuracy of the SNN when no noise is applied. More specifically, more than 20\% drop is noticed on the \textit{MF} with $T=25$, and lower differences for the other filters. However, in the presence of noise, the SNN becomes much more robust when the filter is applied. For example, when considering normal noise with a magnitude of 0.55, the \textit{BAF} with $s=1$ and $t=5$ contributes to 64\% accuracy improvement (see pointer~\rpoint{2}). On the other hand, \textit{BAFs} with $s~\geq~2$ do not increase the accuracy much, compared to the unfiltered SNN. Moreover, \textit{MFs} with $T~\geq~100$ work even better than the BAFs in the presence of large perturbations. Indeed, the perturbations with magnitude of 1.0 are filtered out relatively well by the \textit{MFs} with large $T$ (see pointer~\rpoint{3}), while, for the same noise magnitude, both the \textit{MFs} with $T~\leq~50$ and the \textit{BAF} with $s=1$ and $t=5$ achieve an accuracy of only $\approx$33-34\% (see pointer~\rpoint{4}). The key message learnt from the above case study is that the noise filters for DVS can potentially restore a large portion of accuracy that would have been dropped due to the perturbations. Therefore, this motivates us to employ such filters as defense methods against adversarial attacks.


\subsection{Our Novel Contributions}

\begin{itemize}[leftmargin=*]
    \item We propose \textit{DVS-Attacks}, a set of different adversarial attack methodologies generating perturbations for DVS signals; (\textbf{Section~\ref{sec:attacks}}). As per our knowledge, \textit{these are the first proposed attack algorithms for event-based neuromorphic systems}.
    \item In particular, the \textit{MF-Aware Dash Attack} is specifically designed to be resistant against the \textit{Mask Filter} defense, by generating perturbations only on a limited set of frames; (\textbf{Section~\ref{sec;MF-Aware_dash_attack}}).
    \item The experimental results on the DVS-Gesture and NMNIST datasets show that all the attacks are successful when no filter is applied. Moreover, the noise filters cannot fully defend against the \textit{DVS-Attacks}, which represent a serious security threat for SNN-based neuromorphic systems; (\textbf{Section~\ref{sec:results}}).
    \item For reproducible research, we released the source code of all the proposed \textit{DVS-Attacks} methodologies, and filters for DVS-based SNNs at \url{https://github.com/albertomarchisio/DVS-Attacks}.
\end{itemize}

Before proceeding to the technical details, \textbf{Section~\ref{sec:background}} presents an overview of SNNs, noise filters for DVS signal, adversarial attacks, and security threats for SNNs. Moreover, \textbf{Section~\ref{sec:threat_model}} discusses the threat model employed in this work.

\section{Background and Related Work}
\label{sec:background}

\subsection{Spiking Neural Networks (SNNs)}

SNNs are considered as the third generation neural networks~\cite{Maas1997ThirdGenerationSNN}. Compared to the traditional DNNs, they exhibit better biological plausibility~\cite{Kasinski2011IntroSNN} and high resilience~\cite{Schuman2020ResilienceSNN}\cite{Putra2021QSpiNN}\cite{Putra2021SparkXD} compared to the traditional DNNs~\cite{Marchisio2019DL4EC}\cite{Capra2020Updated}. Another key advantage of SNNs over the traditional DNNs is their improved energy-efficiency~\cite{Putra2020FSpiNN}\cite{Putra2021SpikeDyn} when implemented on Neuromorphic chips like Intel Loihi~\cite{Davies2018Loihi} or IBM TrueNorth~\cite{Merolla2014Truenorth}. Moreover, the recent development of DVS sensors~\cite{Lichtsteiner2006DVS} has further reduced the energy requirements of the complete system~\cite{Massa2020EfficientSNN}\cite{Viale2021CarSNN}.

\begin{figure}[h]
	\centering
	\includegraphics[width=\linewidth]{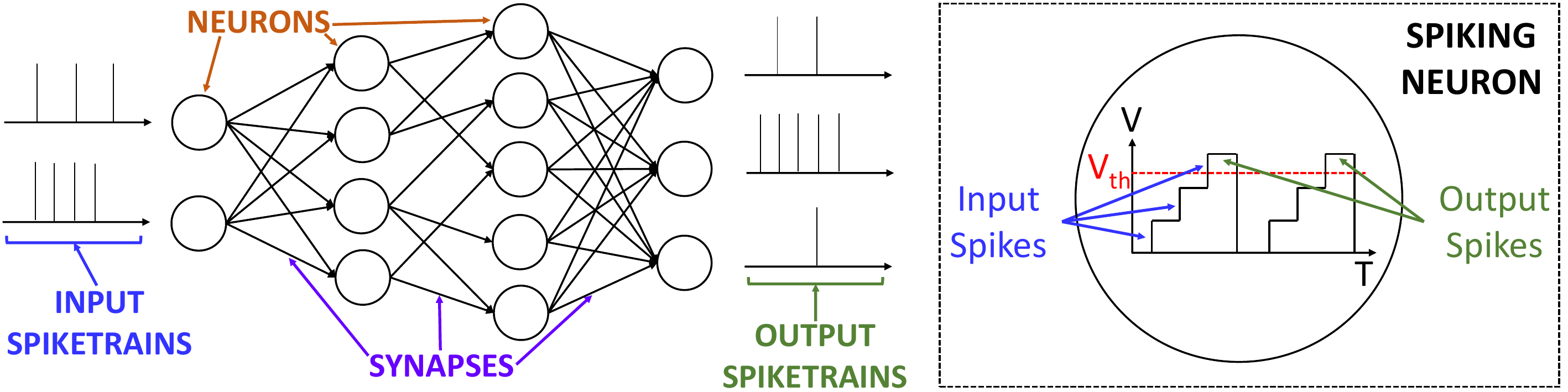}
	\caption{Overview of an SNN's functionality, focusing on the evolution over time of the membrane potential of a spiking neuron.}
	\label{fig:SNN}
\end{figure}

In SNNs, the input is encoded using spikes, which propagate to the output through neurons and synapses. In a Leaky-Integrate-and-Fire (LIF) neuron, which is the most commonly adopted spiking neuron model, each input spike contributes to increasing the neuron membrane potential $V$ over time. As shown in Fig.~\ref{fig:SNN}, when $V$ overcomes a threshold $V_{th}$, an output spike is released by the neuron, and propagated to the neurons of the following layer.

\textbf{Event-based cameras}~\cite{Lichtsteiner2006DVS} are the new generations of bio-inspired sensors for the acquisition of visual information, directly related to the light variations in the scene. Instead of recording frames with a precise timing, the DVS cameras work asynchronously, recording only positive and negative brightness variations in the scene. Each event is encoded with four components $(x,y,p,t)$, which represent the x-coordinate, the y-coordinate, the polarity, and the timestamp, respectively. Compared to classical frame-based image sensors, the event-based sensors consume significantly lower power, since the events are recorded only when a brightness variation in the scene is detected. This means that, in the absence of light changes, no information is recorded, leading close to zero power consumption. \textit{Hence, DVS sensors can be efficiently deployed at the edge and directly coupled to neuromorphic hardware for low-power SNN-based applications.}

\subsection{Noise Filters for Dynamic Vision Sensors}
\label{subsec:noise_filters}

DVS sensors are mainly affected by background activity noise, caused by thermal noise and junction leakage current~\cite{Nozaki2017ParasiticDVS}. When the DVS is stimulated, a neighborhood of pixels is usually active at the same time, generating events. Therefore, the real events show a higher spatio-temporal correlation than the noise-related events. This empirical observation is exploited for generating the \textbf{Background Activity Filter (BAF)}~\cite{LinaresBarranco2019FilterDVS}. The events are associated with a spatio-temporal neighborhood, within which the correlation between them is calculated. If the correlation is lower than a certain threshold, the events are likely due to noise and thus are filtered out; otherwise they are kept. 
The procedure is reported in Algorithm~\ref{alg:BAF}, where $S$ and $T$ are the only parameters of the filter and are used to set the dimensions of the spatio-temporal neighborhood. The larger $S$ and $T$ are, the lower the number of events are filtered out. The decision of the filter is made by the comparison between $t_e - M[x_e][y_e]$ and $T$ (lines 15-16 of Algorithm~\ref{alg:BAF}). If the first term is lower, then the event is filtered out.

\begin{figure}[h]
\begin{algorithm}[H]
\caption{\textbf{:} \textit{Background Activity Filter} for event-based sensors.}
\label{alg:BAF}
\begin{small}
\begin{algorithmic}[1]
\vspace{-5pt}
\STATE Being $E$ a list of events of the form $(x,y,p,t)$
\STATE Being $(x_e,y_e,p_e,t_e)$ the x-coordinate, the y-coordinate, the polarity and the timestamp of the event $e$ respectively
\STATE Being $M$ a $128 \times 128$ matrix
\STATE Being S and T the spatial and temporal filter's parameters
\STATE Initialize $M$ to zero
\STATE Order $E$ from the oldest to the newest event
\FOR  {$e$ in $E$}
    \FOR {$i$ in ($x_e-S$,$x_e+S$)}
        \FOR {$j$ in ($y_e-S$, $y_e+S$)}
            \IF { not ($i == x_e$ and $j==y_e$)}
                \STATE $M[i][j]= t_e$
            \ENDIF
        \ENDFOR
    \ENDFOR
    \IF {$t_e - M[x_e][y_e]>T$}
        \STATE Remove $e$ from $E$
    \ENDIF
\ENDFOR
\end{algorithmic}
\end{small}
\end{algorithm}
\end{figure}

Another type of scenario in which spontaneous noise activity is generated on the pixels which have low temporal contrast. In this case, a \textbf{Mask Filter (MF)} is required to filter-out such noise~\cite{LinaresBarranco2019FilterDVS}. The procedure reported in Algorithm~\ref{alg:MF} shows that, compared to the \textit{BAF}, the \textit{MF} has only the temporal parameter $T$. If the activity of a pixel exceeds $T$, the mask is activated (lines~14-15 of Algorithm~\ref{alg:MF}). After all the pixel coordinates of the mask are set, each event generated on a coordinate in which the mask is active is removed (lines~20-21). \textit{Both the \textit{BAF} and \textit{MF} have been implemented and evaluated in the presence of intrinsic and parasitic noise of DVS sensors, while their application as a defense mechanism against adversarial attacks is still unexplored.}

\begin{figure}[h]
\begin{algorithm}[H]
\caption{\textbf{:} \textit{Mask Filter} for event-based sensors.}
\label{alg:MF}
\begin{small}
\begin{algorithmic}[1]
\vspace{-5pt}
\STATE Being $E$ a list of events of the form $(x,y,p,t)$,
\STATE Being $(x_e,y_e,p_e,t_e)$ the x-coordinate, the y-coordinate, the polarity and the timestamp of the event $e$ respectively,
\STATE Being $M$ a $N \times N$ matrix, where $N$ is the size of the frames,
\STATE Being $activity$ a $N \times N$ matrix, representing the number of event produced by each pixel,
\STATE Being T, the temporal threshold passed to the filter as a parameter,
\STATE Initialize $activity$ to zero
\FOR {x in range(N)}
	\FOR {y in range(N)}
		\FOR {e in E}
			\IF {$(x,y) == (x_e,y_e)$ }
				\STATE $activity[x][y]+= 1$
			\ENDIF
		\ENDFOR
		\IF {$ activity[x][y] > T $ }
			\STATE	 $M[x][y] = 1$
		\ENDIF
	\ENDFOR
\ENDFOR
\FOR {e in E}
	\IF {$M[x_e][y_e]== 1$ }
		\STATE Remove e from E
	\ENDIF
\ENDFOR
\end{algorithmic}
\end{small}
\end{algorithm}
\end{figure}

\begin{figure*}[h]
    \centering
    \includegraphics[width=\linewidth]{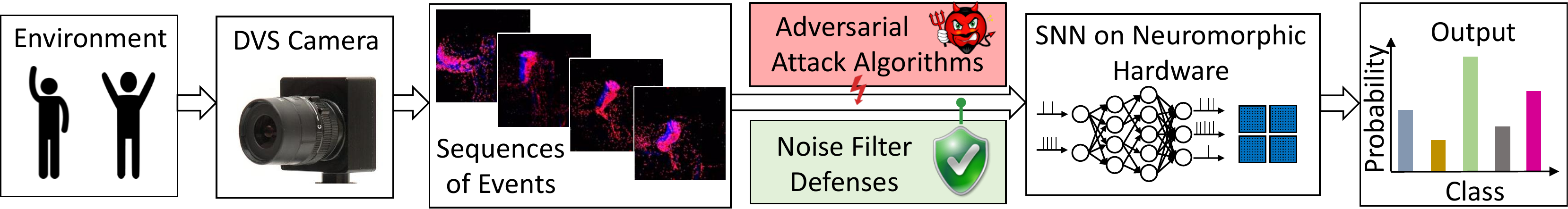}
    \caption{Threat model considered in this work. Different types of adversarial attacks are considered, and different types of noise filters are applied as defense.}
    \label{fig:threat_model}
\end{figure*}

\subsection{Adversarial Attacks and Security Threats for SNNs in the Spatio-Temporal Domain}

Currently, adversarial attacks are deployed on a wide range of deep learning applications~\cite{Shafique2020RobustMLDnT}. They represent a serious threat for safety-critical applications, like surveillance, medicine, and autonomous driving~\cite{Cheng2018SafetyCriticalDNN}. The objective of a successful attack is to generate small imperceptible perturbations to fool the network. Recently, adversarial attacks for SNNs have been explored, working in black-box~\cite{Marchisio2019SNNAttack} and white-box settings~\cite{Bagheri2018AdvTrainingSNN}.
Sharmin et al.~\cite{Sharmin2019ACA} proposed a methodology to attack (non-spiking) DNNs, and then the adversarial examples mislead the equivalent converted SNNs. 
Liang et al.~\cite{Liang2020ExploringAA} proposed a gradient-based adversarial attack methodology for SNNs. Venceslai et al.~\cite{Venceslai2020NeuroAttack} proposed a methodology to attack SNNs through bit-flips triggered by adversarial perturbations. Towards adversarial robustness, recent works demonstrated that SNNs are inherently more robust than DNNs, due to the effect of effects of discrete
input encoding, non-linear activations, and structural parameters~\cite{Sharmin2020InherentAdvSNN}\cite{ElAllami2021SecuringSNNInherent}. \textit{However, none of these previous works analyze attacks or defenses on frames of events, coming from DVS cameras.}

While in adversarial attack algorithms for images the perturbations are only added in the spatial domain, an attack on the DVS signal must introduce perturbations also in the temporal domain. As per our knowledge, there are no existing adversarial attack methodologies for event-based cameras coupled with SNN processing hardware. Some related works can be found in the field of attacks on video signals, i.e., sequences of events. State-of-the-art adversarial attacks on videos include, among others, sparse adversarial perturbations~\cite{Wei2019SparseAP}, where only a small subset of frames are perturbed. In this way, the attack is stealthy, because only a few frames are perturbed, and effective, due to the temporal interaction between consecutive frames. Another state-of-the-art method is represented by the adversarial framing~\cite{Zajac2019AdversarialFraming}, where the perturbation is added to the border of the frames and the misclassification is achieved. \textit{However, frames of events cannot be treated as videos, since the latter contain the information of pixel intensities for every frame, while do not contain other types of information, such as the polarity. Hence, the adversarial attacks for videos cannot be directly applied to DVS signals.}

\section{Threat Model}
\label{sec:threat_model}

The system that we use in our experiments is composed of a DVS camera, for recording the scenes of the environment as sequences of events, and a given SNN implemented onto the neuromorphic hardware. As shown in Fig.~\ref{fig:threat_model}, the adversarial attacks and noise filters for defense are located at the input of the SNN, and have access to modify the sequences of events. We conduct several experiments with different combinations of attacks and defenses. The noise filters described in Section~\ref{subsec:noise_filters} have been employed as defense methods. For the combinations in which both attacks and defenses are present in the system, the modifications generated by the attack are applied before the filter operation. In this way, the filter has the ability to filter out any events that have been generated or modified by the attack algorithm, thus aiming at making the defense stronger. The detailed description of the adversarial attack methodologies is discussed in the following Section~\ref{sec:attacks}.

\section{Our Proposed DVS-Attacks Methodologies}
\label{sec:attacks}

\subsection{Sparse Attack}
The proposed \textit{Sparse Attack} is an iterative algorithm, which progressively updates the perturbation values based on the loss function (lines~6-12 of Algorithm~\ref{alg:Sparse Attack}), for each frame series of the dataset $D$. A mask $M$ determines in which subset of the frames of events the perturbation should be added (line~7). Then, the output probability and the respective loss, obtained in the presence of the perturbation, are respectively computed in lines~9 and~10. Finally, the perturbation values are updated based on the gradients of the inputs w.r.t. the loss (line~11).

\begin{figure}[h]
\begin{algorithm}[H]
\caption{\textbf{:} \textit{Sparse Attack} Methodology.}
\label{alg:Sparse Attack}
\begin{small}
\begin{algorithmic}[1]
\vspace{-5pt}
\STATE Being $M$ a mask able to select only certain frames
\STATE Being $D$ an event-based dataset
\STATE Being $P$ a perturbation to be added to the images
\STATE Being $prob$ the output probability of a certain class
\FOR  {$d$ in $D$}
    \FOR {$i$ in $max\_iteration$}
    \STATE Add $P$ to $d$ only on the frames selected by $M$
    \STATE Calculate the prevision on the perturbed input
    \STATE Extract $prob$ for the actual class of $d$
    \STATE Update the loss value as $loss=-log (1- prob$)
    \STATE Calculate the gradients and update $P$
    \ENDFOR
\ENDFOR
\end{algorithmic}
\end{small}
\end{algorithm}
\end{figure}

\subsection{Frame Attack}
The \textit{Frame Attack} is a simple yet effective attack methodology, which consists of adding a frame around the sample (lines~6-8 of Algorithm~\ref{alg:Frame Attack}). It does not require any expensive calculations, because the same perturbation (which coincides with the frame) is added to all the samples.
In a dataset made of large images, such as the DVS-Gesture ($128~\times~128$) it is also not so easy to spot, while with the perturbations on the NMNIST dataset ($34~\times~34$) result more evident.
One drawback is due to the overhead added to the samples in terms of events. In fact, since the attack targets every pixel of the boundary, for every frame, the number of events dramatically increases. Therefore, the size of the samples and the inference latency to process the events with the SNN and the filters increase as well.
\begin{figure}[ht]
\begin{algorithm}[H]
\caption{\textbf{:} \textit{Frame Attack} Methodology.}
\label{alg:Frame Attack}
\begin{small}
\begin{algorithmic}[1]
\vspace{-5pt}
\STATE Being $D$ an event-based dataset
\STATE Being $d \subset D $  a $(C \times N \times N \times T )$ tensor, where C represents the channels, N represents the frame dimensions, and T the sample duration
\FOR {d in D}
	\FOR {x in range(N)} 
		\FOR {y in range(N)}
			\IF {$x==0$ \OR $x==N-1$ \OR $y==0$ \OR $y==N-1$ }
				\STATE $d[:,x,y,:] = 1$
			\ENDIF
		\ENDFOR
	\ENDFOR
\ENDFOR
\end{algorithmic}
\end{small}
\end{algorithm}
\end{figure}

\subsection{Corner Attack}

The \textit{Corner Attack}, as the name suggests, targets the corner of the images.
It starts by modifying only two pixels at the top-left corner (lines~10-11 of Algorithm~\ref{alg:Corner Attack}) and then, if it is not successful in fooling the SNN (line~16), it moves to the other corners.
If some samples remain correctly classified, after hitting all 4 corners, the size of the perturbation increases and the algorithm resumes from the first corner.
Before the updating phase, both when it changes corner or when it increase its size, the attack is applied to every sample in the dataset that was not yet corrupted. In this way, as the algorithm proceeds, the number of samples reduces and the the process is sped up. The main feature of this attack is that not all the samples are modified by the same amount of perturbation. For example, while the majority of the samples are misinterpreted by the SNN after few iterations, other samples are perturbed for longer time, thus making the attack easier to spot.

\begin{figure}[ht]
\begin{algorithm}[H]
\caption{\textbf{:} \textit{Corner Attack} Methodology.}
\label{alg:Corner Attack}
\begin{small}
\begin{algorithmic}[1]
\vspace{-5pt}
\STATE Being $D$ an event-based dataset made of $(C \times N \times N \times T )$ tensors, where C represents the number of channels, N the size, and T the duration of the sample
\STATE $S$ is a list of the samples that compose $D$
\STATE $x = 0 $
\STATE $y = 2 $
\STATE $left= True$
\WHILE {$S$ is not empty}
	\FOR {s in S}
		\FOR {i in range(N)}
			\FOR {j in range(N)}
				\IF {$i==x$ \AND ($left$ \AND $j<y$ \OR $\overline{left}$ \AND $j\geq N-y-1)$}
					\STATE $s[:,i,j,:] = 1$
				\ENDIF
			\ENDFOR
		\ENDFOR
		\STATE The perturbed sample s is fed to the SNN, which produces a prediction P
		\IF {P is incorrect}
			\STATE Remove s from S
		\ENDIF
	\ENDFOR
	\IF {$x == 0$}
		\STATE $x = N-1$
	\ELSE
		\STATE $left = left$ \XOR $1$
		\STATE $x = 0$
		\IF {$\overline{left}$}
			\STATE $y = y+1$
		\ENDIF
	\ENDIF
\ENDWHILE
\end{algorithmic}
\end{small}
\end{algorithm}
\end{figure}

\subsection{Dash Attack}

The \textit{Dash Attack} methodology is designed taking inspiration from the \textit{Corner Attack}. Indeed, the two algorithms are quite similar. The main difference is that in the \textit{Dash Attack} only two pixels are targeted every time. The main structure of the algorithm is the same as for the \textit{Corner Attack}, as the \textit{Dash Attack} starts by targeting the top-left corner and by modifying the first two pixels. Moreover, the $x,y$ coordinates are updated, in order for the attack to hit only two consecutive pixels (see lines~19-29 of Algorithm~\ref{alg:Dash Attack}). Hence, this attack results to be very difficult to spot, and the introduced perturbations do not cause a large overhead of events on the samples. Moreover, all the samples under the \textit{Dash Attack} are modified by the same amount of perturbations.

\begin{figure}[ht]
\begin{algorithm}[H] \label{dash_alg}
\caption{\textbf{:} \textit{Dash Attack} Methodology.}
\label{alg:Dash Attack}
\begin{small}
\begin{algorithmic}[1]
\vspace{-5pt}
\STATE Being $D$ an event-based dataset made of $(C \times N \times N \times T )$ tensors, where C represents the number of channels, N the size, and T the duration of the sample
\STATE $S$ is a list of the samples that compose $D$
\STATE $x_{min}=0$, $x=0 $, $y=2$
\STATE $left= True$
\WHILE {$S$ is not empty}
	\FOR {s in S}
		\FOR {i in range(N)}
			\FOR {j in range(N)}
				\IF {$i==x$ \AND ($left$ \AND ($j==y$ \OR $j== y-1$) \OR $\overline{left}$ \AND ($j==N-y$ \OR $j==N-y+1))$}
					\STATE $s[:,i,j,:] = 1$
				\ENDIF
			\ENDFOR
		\ENDFOR
		\STATE The perturbed sample s is fed to the SNN, which produces a prediction P
		\IF {P is incorrect}
			\STATE Remove s from S
		\ENDIF
	\ENDFOR
	\IF {$x == x_{min}$}
		\STATE $x = N-x_{min}-1$
	\ELSE
		\STATE $left = left$ \XOR $1$
		\STATE $x = x_{min}$
		\IF {$\overline{left}$}
			\STATE $y = y+1$
		\ENDIF
	\ENDIF
	\IF {$y>N/2$}
	    \STATE $x_{min}= x_{min}+1$
	\ENDIF

\ENDWHILE
\end{algorithmic}
\end{small}
\end{algorithm}
\end{figure}

\subsection{MF-Aware Dash Attack}
\label{sec;MF-Aware_dash_attack}

The main issue of the above-discussed attacks, as will be demonstrated in Section~\ref{sec:results}, is their intrinsic weakness against the \textit{MF}. In fact, they targeted both channels (`on' and `off' events) of the same pixels for all the duration of the sample. This leads to a clear distinction between the pixels affected by the attack and those that are not. In fact, the number of events produced by the targeted pixels is significantly higher than the events associated to the other pixel coordinates that were not hit by the attack. In addition, we have to consider the fact that the proposed attacks mainly focus on the boundaries of the images, thus they do not tend to overlap with useful information. In other words, in the datasets that we used the subject is typically centered. Hence, by hitting the perimeter or the corners, the risk of superimposing adversarial noise to the main subject is low. 
These considerations explain why the \textit{MF} is successful for restoring the original SNN accuracy. Indeed, the targets are easily identifiable given their high number of events, and the filter does not remove useful information, since modifications are mainly conducted at the edge of the image.

Based on these premises, we have designed an attack aiming at being resistant to the \textit{MF}, which we call \textit{MF-Aware Dash Attack}. It receives as a parameter $th$, which is correlated to the $T$ parameter of the \textit{MF} (recall from Algorithm~\ref{alg:MF}), and it uses it to set a limit on the number of frames that can be changed for each pixel (line~11 of Algorithm~\ref{alg:MF-Aware Dash Attack}). Therefore, the algorithm targets a couple of pixels to be perturbed, as in case of the \textit{Dash Attack}. However, after modifying $th$ frames, it moves to the following ones (lines~16-18). The visual effect generated by the \textit{MF-Aware Dash Attack} is that of a dash advancing along a line. The smaller the parameter $th$ is, the faster will the dash seem moving along the image.

\begin{figure}[ht]
\begin{algorithm}[H] \label{dash_sp_alg}
\caption{\textbf{:} \textit{MF-Aware Dash Attack} Methodology.}
\label{alg:MF-Aware Dash Attack}
\begin{small}
\begin{algorithmic}[1]
\vspace{-5pt}
\STATE Being $D$ an event-based Dataset made of $(2 \times N \times N \times T )$ tensors, where N represents the frame dimensions, and T the sample duration
\STATE $S$ is a list of the samples that compose $D$
\STATE $th$ is a parameter associated the activity threshold of the \textit{MF}  
\STATE $x = 0 $ , $y_0 = 2 $, $left= True$
\WHILE {$S$ is not empty}
	\FOR {s in S}
		\STATE $th = th_0$, $y=y_0$
		\FOR {t in T}
			\FOR {i in range(N)}
				\FOR {j in range(N)}
					\IF {$i==x$ \AND $t<th$ \AND ($left$ \AND ($j==y$ \OR $j== y-1$) \OR $\overline{left}$ \AND ($j==N-y$ \OR $j==N-y+1))$}
						\STATE $s[0,i,j,t] = 1$
					\ENDIF
				\ENDFOR
			\ENDFOR
			\IF{$t == th$}
				\STATE $th = th+ th_0$ , $y=y+2$
			\ENDIF
		\ENDFOR
	
		\STATE The perturbed sample s is fed to the SNN, which produces a prediction P
		\IF {P is incorrect}
			\STATE Remove s from S
		\ENDIF
	\ENDFOR
	\IF {$x == 0$}
		\STATE $x = N-1$
	\ELSE
		\STATE $left = left$ \XOR $1$ , $x = 0$
		\IF {$\overline{left}$}
			\STATE $y_0 = y_0+1$
		\ENDIF
	\ENDIF
\ENDWHILE
\end{algorithmic}
\end{small}
\end{algorithm}
\end{figure}

\section{Evaluation of the DVS-Attacks in the Presence of Noise Filters}
\label{sec:results}

\subsection{Experimental Setup}

\begin{figure}[h]
    \includegraphics[width=\linewidth]{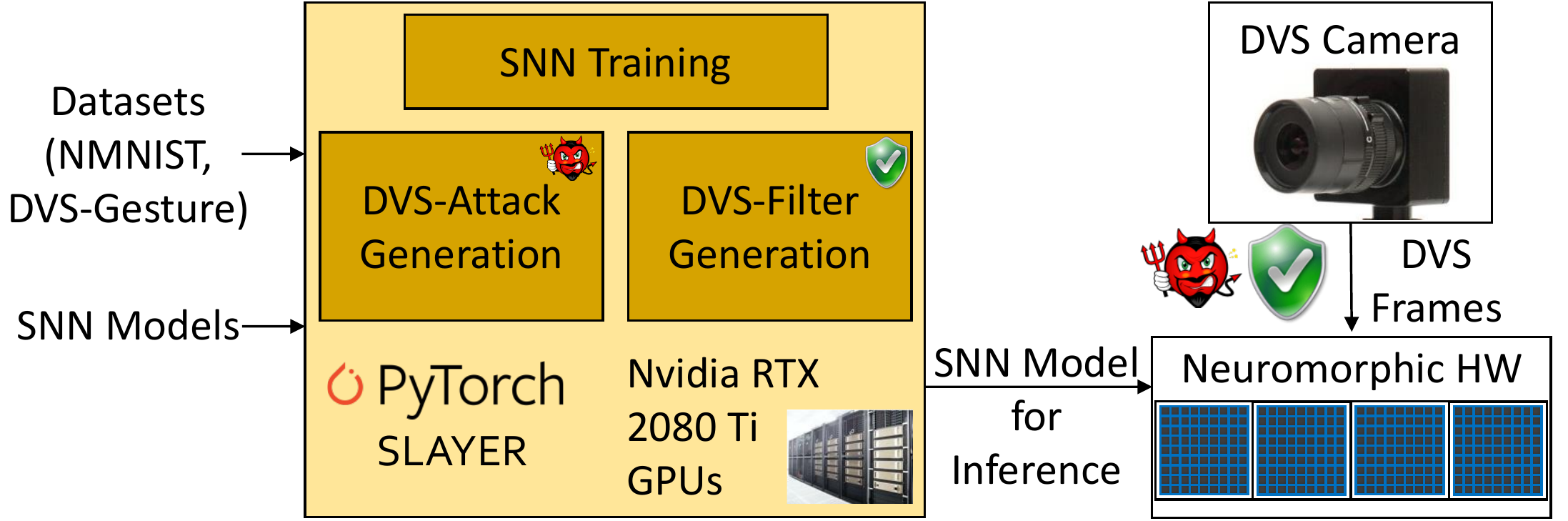}
    \caption{Experimental setup, tool-flow, and integration with the system.}
    \label{fig:exp_setup}
\end{figure}

We conducted experiments on two datasets, the DVS-gesture~\cite{Amir2017DVSgesture} and the NMNIST~\cite{Orchard2015NMNIST}. The former is a collection of of 1077 samples for training and 264 for testing, divided into 11 classes, while the latter is a spiking version of the original frame-based MNIST dataset~\cite{LeCun1998MNIST}. It is generated by an ATIS event-based sensor~\cite{Posch2011ATIS} that is moved while capturing the MNIST images projected on a LCD screen. It consists of 60,000 training and 10,000 testing samples. 
As classifier for the DVS-gesture dataset, we employed the 4-layer SNN as described in~\cite{Shrestha2018SLAYER}, with two convolutional layers and two fully-connected layers, and trained it for 625 epochs with the SLAYER backpropagation method~\cite{Shrestha2018SLAYER}, using a batch size of 4 and learning rate equal to 0.01. We measured a test accuracy of 92.04\% on clean inputs.
As classifier for the NMNIST dataset, we employed a multilayer perceptron with two fully-connected layers~~\cite{Shrestha2018SLAYER}, trained for 350 epochs with the SLAYER backpropagation method~\cite{Shrestha2018SLAYER}, using a batch size of 4 and learning rate equal to 0.01. The test accuracy on clean inputs is 95\%.
We implemented the SNNs on a DL-workstation with two Nvidia GeForce RTX 2080 Ti GPUs, using the PyTorch framework~\cite{pytorch}. We also implemented the adversarial attack algorithms and the noise filters in PyTorch. The experimental setup and tool-flow in a real-world setting is shown in Fig.~\ref{fig:exp_setup}.

\subsection{Results for the Sparse Attack}

\begin{figure*}[h]
    \includegraphics[width=\linewidth]{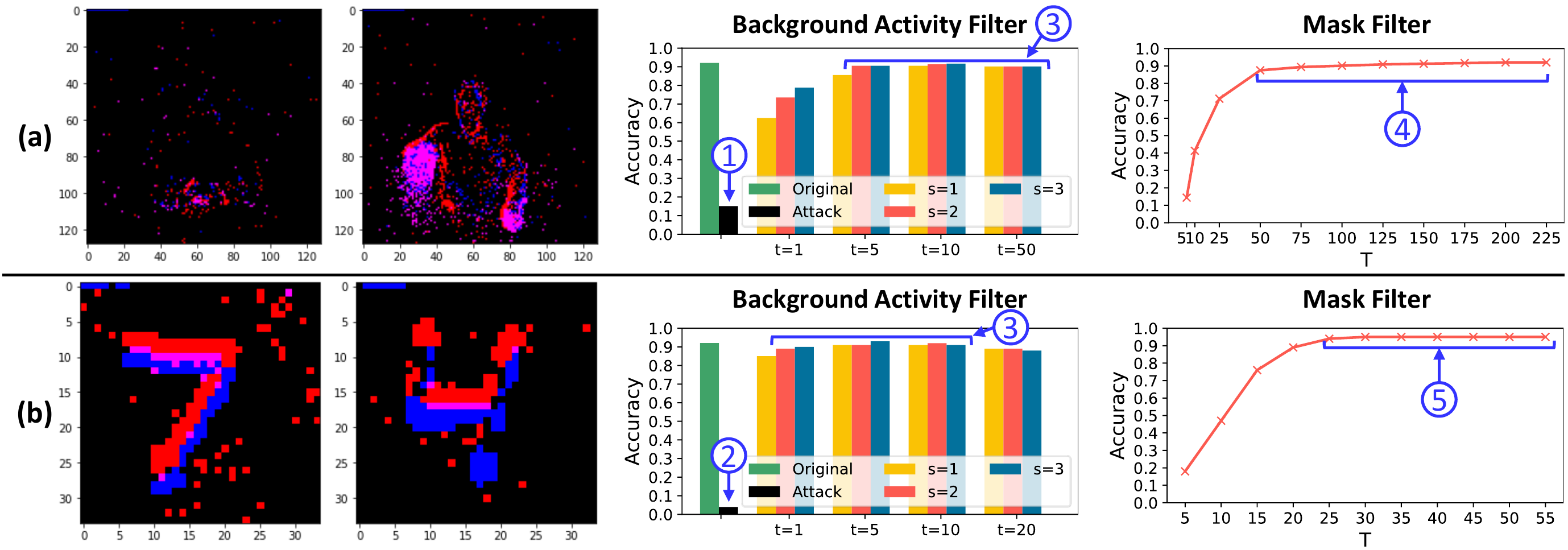}
    \caption{Evaluation of the \textit{Sparse Attack}: frame samples and accuracy when the \textit{BAF} and \textit{MF} are applied, for (a) DVS-Gesture and (b) NMNIST.}
    \label{fig:sparse_results}
\end{figure*}

The \textit{Sparse Attack} on DVS frames is successful on both benchmarks, as the accuracy is drastically decreased to 15.15\% for the DVS-Gesture dataset (see pointer~\rpoint{1} in Fig.~\ref{fig:sparse_results}), and to 4\% for the NMNIST dataset (see pointer~\rpoint{2}). By looking at the adversarial examples reported at the left side of Fig.~\ref{fig:sparse_results}, no significant perturbations are perceived, thus making the \textit{Sparse Attack} stealthy. However, the accuracy can be easily restored using a noise filter. When the \textit{BAF} filter is employed, for a wide range of the $(s,t)$ parameters the SNNs' accuracy overcomes 90\% (see pointers~\rpoint{3}). When the \textit{MF} is used, a low $T$ does not protect well against the \textit{Sparse Attack}, but when $T~\geq~50$ for the DVS-Gesture dataset (see pointer~\rpoint{4}) and when $T~\geq~25$ for the NMNIST dataset (see pointer~\rpoint{5}), high robustness is achieved.

\subsection{Results for the Frame Attack}

\begin{figure*}[h]
    \includegraphics[width=\linewidth]{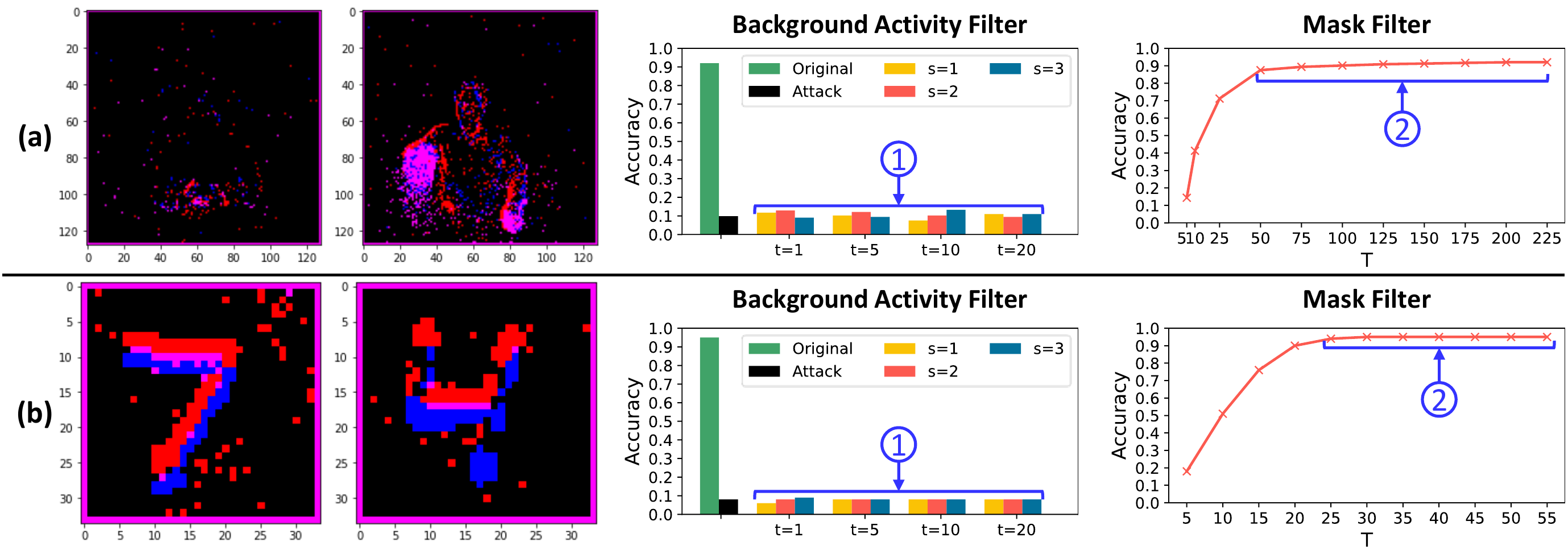}
    \caption{Evaluation of the \textit{Frame Attack}: frame samples and accuracy when the \textit{BAF} and \textit{MF} are applied, for (a) DVS-Gesture and (b) NMNIST.}
    \label{fig:frame_results}
\end{figure*}

The results for the experiments conducted on the \textit{Frame Attack} are reported in Fig.~\ref{fig:frame_results}. As expected, the perturbations are perceivable in the form of a line added to the border of the visualized shot. This feature is much more accentuated on the NMNIST dataset, where the resolution is of $34~\times~34$ pixels, while the perturbations are less distinguishable on the $128~\times~128$ examples of the DVS-Gesture dataset. The accuracy under attack drops to 9.85\% and 8\% for the two datasets, respectively. However, the \textit{BAF} does not work well as a defense against the \textit{Frame Attack}. As highlighted by pointers~\rpoint{1} in Fig.~\ref{fig:frame_results}, there exist no combinations of the $(s,t)$ parameters of the \textit{BAF} for which the SNNs' accuracy significantly increases. Indeed, the accuracy difference compared to the attack without filter is relatively low. On the other hand, the \textit{MF} results to be a successful defense, because the SNNs' accuracy is high for large values of $T$ (see pointer~\rpoint{2}).

\subsection{Results for the Corner Attack}

\begin{figure*}[h]
    \includegraphics[width=\linewidth]{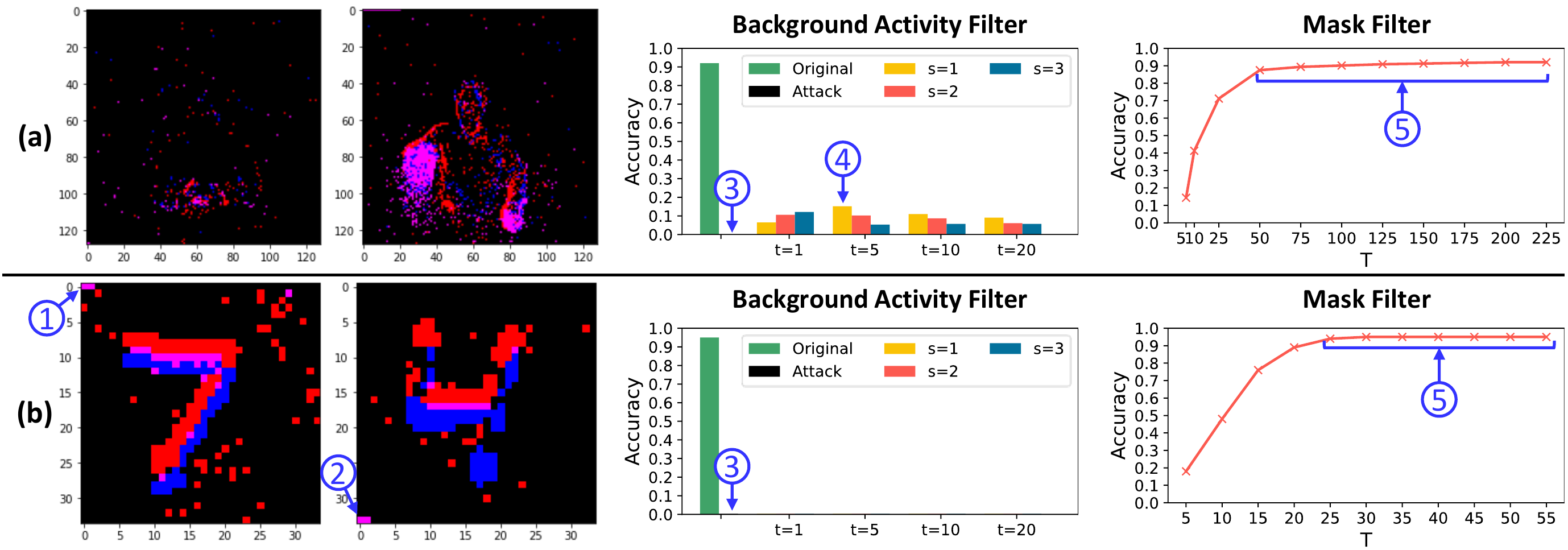}
    \caption{Evaluation of the \textit{Corner Attack}: frame samples and accuracy when the \textit{BAF} and \textit{MF} are applied, for (a) DVS-Gesture and (b) NMNIST.}
    \label{fig:corner_results}
\end{figure*}

The \textit{Corner Attack} is visibly stealthier than the \textit{Frame Attack}. Indeed, the perturbations are only added in the corner of the images. For example, the perturbation is noticeable in the top-left corner of the first example of the NMNIST dataset (see pointer~\rpoint{1} in Fig.~\ref{fig:corner_results}), or in the bottom-left corner of the second example (see pointer~\rpoint{2}). Moreover, the SNNs are completely fooled by the \textit{Corner Attack}, since the accuracy without noise drops to 0\% (see pointers~\rpoint{3}). The \textit{BAF} works relatively better for the DVS-Gesture dataset, compared to the MNIST dataset. However, the accuracy in the presence of the \textit{BAF} filter as defense remains very low. The peak reached with $s=1$ and $t=5$ has an accuracy of only 15.15\% for the SNN on the DVS-Gesture dataset (pointer~\rpoint{4}). Similarly to the \textit{Frame Attack}, also the \textit{Corner Attack} can be successfully mitigated when the \textit{MF} with large $T$ is applied (see pointers~\rpoint{5}).

\subsection{Results for the Dash Attack}

\begin{figure*}[h]
    \includegraphics[width=\linewidth]{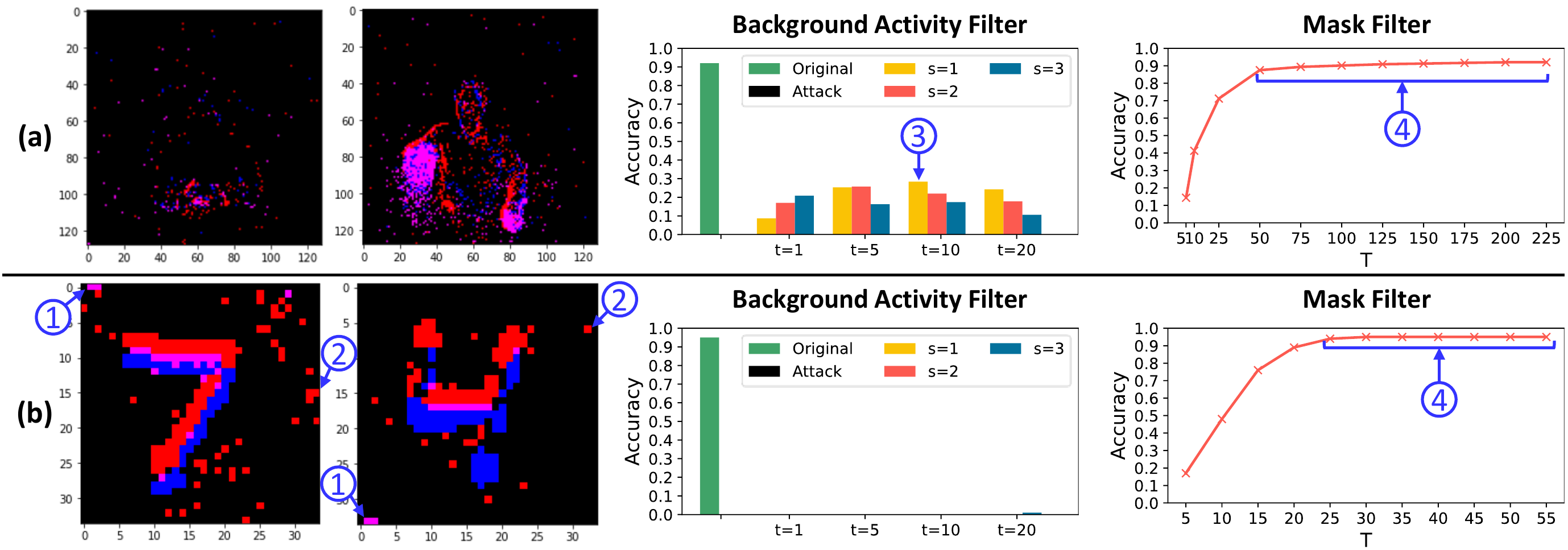}
    \caption{Evaluation of the \textit{Dash Attack}: frame samples and accuracy when the \textit{BAF} and \textit{MF} are applied, for (a) DVS-Gesture and (b) NMNIST.}
    \label{fig:dash_results}
\end{figure*}

The \textit{Dash Attack} performs in a similar way as the \textit{Corner Attack}, but the perturbations are not strictly confined in a corner. In this way, the perturbations introduced by the attack result very similar to the inherent background noise generated by the DVS camera recording the events. For instance, the attack perturbations on the examples for the NMNIST dataset (see pointers~\rpoint{1} in Fig.~\ref{fig:dash_results}) might be confused with the inherent background noise (see pointers~\rpoint{2}). Compared to the \textit{Corner Attack}, while the accuracy of the SNNs under the \textit{Dash Attack} without filter drops to 0\%, the \textit{BAF} defense produces a slightly higher SNN accuracy for the DVS-Gesture dataset. However, the accuracy peak of 28.41\% (see pointer~\rpoint{3}), obtained in the presence of the \textit{BAF} with $s=1$ and $t=10$, remains too low to consider the \textit{BAF} as a good defense method against the \textit{Dash Attack}. Once again, a good defense for robust SNNs is achieved by the \textit{MF} with large $T$ (see pointers~\rpoint{4}).

\subsection{Results for the MF-Aware Dash Attack}

\begin{figure*}[h]
    \includegraphics[width=\linewidth]{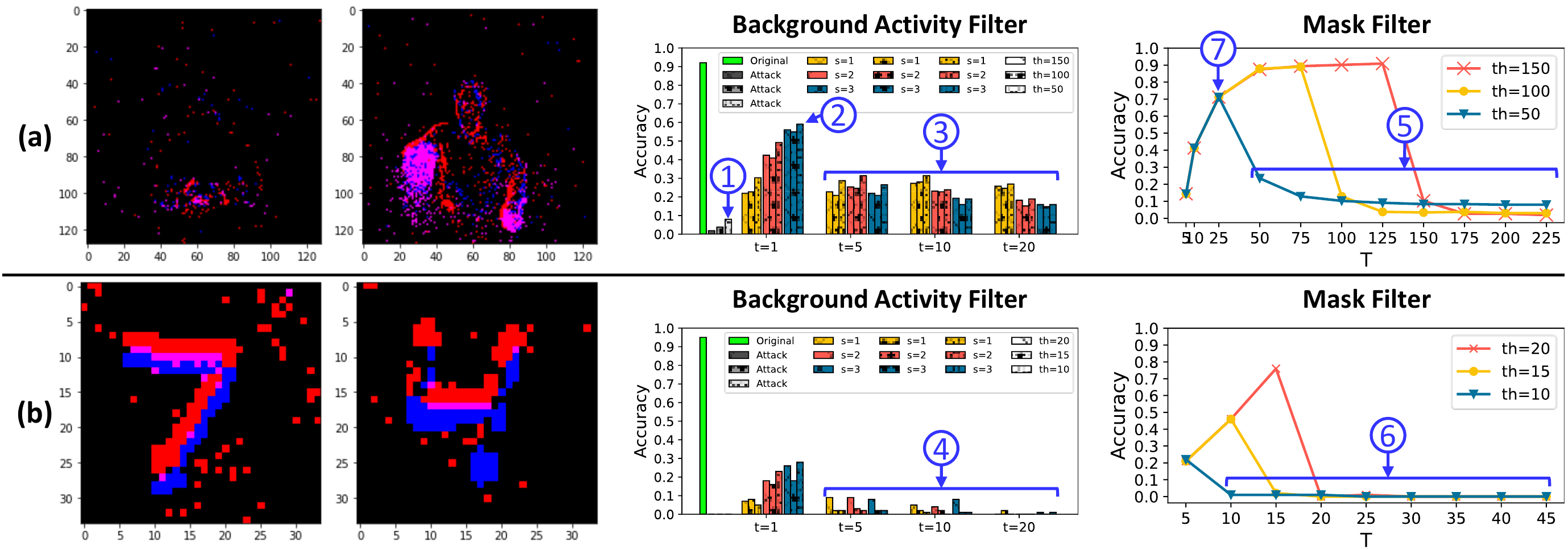}
    \caption{Evaluation of the \textit{MF-Aware Dash Attack}: frame samples and accuracy when the \textit{BAF} and \textit{MF} are applied, for (a) DVS-Gesture and (b) NMNIST. On the left side are reported two adversarial frame samples generated with $th=150$ for the DVS-Gesture dataset, and with $th=20$ for the NMNIST dataset.}
    \label{fig:aware_dash_results}
\end{figure*}

Fig.~\ref{fig:aware_dash_results} shows the results for the experiments conducted on the \textit{MF-Aware Dash Attack}, for different values of the parameter $th$. While the stealthiness of the adversarial examples (Fig.~\ref{fig:aware_dash_results} reports the samples generated with $th=150$ for the DVS-Gesture dataset and $th=20$ for the NMNIST dataset) is similar to the \textit{Corner} and \textit{Dash Attacks}, the behavior of the \textit{MF-Aware Dash Attack} in the presence of noise filters is much different. Moreover, the accuracy of the SNNs under attack without filter are different from 0, reaching up to 7.95\% for $th=50$ on the DVS-Gesture dataset (see pointer~\rpoint{1} in Fig.~\ref{fig:aware_dash_results}). The SNNs defended by the \textit{BAF} show discrete robustness, in particular when $s=3$ and $t=1$. In such scenario, the accuracy reaches 59.09\% when the \textit{MF-Aware Dash Attack} with $th=50$ is applied to the SNN for the DVS-Gesture dataset (see pointer~\rpoint{2}). However, when $t~\geq~5$, the SNN accuracy is lower than 31.44\% for the DVS-Gesture dataset (see pointer~\rpoint{3}) and lower than 13\% for the NMNIST dataset (see pointer~\rpoint{4}). The key advantage compared to the above-discussed attacks resides in the behavior of the \textit{MF-Aware Dash Attack} in the presence of the \textit{MF}. If $T~\geq~th$, the SNN accuracy becomes lower than 23.5\% for the the DVS-Gesture dataset (see pointer~\rpoint{5}) and lower than 2\% for the NMNIST dataset (see pointer~\rpoint{6}). On the contrary, the behavior when $T<th$ is similar to the results obtained for the other attacks. For example, the curve relative to the \textit{MF-Aware Dash Attack} with $th=50$ for the DVS-Gesture dataset achieves 71.21\% accuracy for $T=25$ (see pointer~\rpoint{7}), which is 20.83\% lower than the original SNN accuracy.

\subsection{Key Observations Derived from the Experiments}

By analyzing in more detail the results for the different types of attacks, we can derive the following key observations:

\begin{itemize}[leftmargin=*]
    \item All the attack algorithms belonging to the \textit{DVS-Attacks} set are successful when no filter is applied, since the SNNs' accuracy is significantly decreased.
    \item The \textit{Sparse Attack} is the stealthiest attack, while \textit{Corner}, \textit{Dash} and \textit{MF-Aware Dash Attacks} are sthealtier than the \textit{Frame Attack}.
    \item The \textit{BAF} achieves good defense only for the \textit{Sparse Attack}, while all the other attacks can fool SNNs defended by the \textit{BAF}. Some accuracy is recovered for the \textit{MF-Aware Dash Attack}, but a considerable accuracy loss is measured.
    \item Different $(s,t)$ parameters of the \textit{BAF} need to be evaluated for obtaining the highest accuracy, and the the combinations of these parameters can vary according to different attack algorithms.
    \item The \textit{MF} with large $T$ is a good defense for almost all the attacks, but it does not work well with the \textit{MF-Aware Dash Attack}, since it is an adversarial attack specifically designed for being resistant to the \textit{MF}.
    \item The best \textit{MF-Aware Dash Attack}, that is with $th=50$ for the DVS-Gesture dataset, and with $th=10$ for the NMNIST dataset, can reduce the accuracy by at least 20\% and 65\% for the two datasets, respectively.
\end{itemize}

\section{Conclusion}

In this paper, we designed \textit{DVS-Attacks}, a set of adversarial attack methodologies for SNNs, which introduce the perturbations into the sequences of events. Therefore, they are suitable for neuromorphic systems supplied by DVS cameras. Moreover, two types of noise filters, namely the \textit{Background Activity Filter} and the \textit{Mask Filter}, are applied as defenses. The experimental results show the high success of the attacks, since the SNNs cannot be completely defended by the noise filters. Therefore, they represent critical security threats for SNN-based neuromorphic systems supplied by event-based sensors. We released the source code of the \textit{DVS-Attacks} and noise filters at \url{https://github.com/albertomarchisio/DVS-Attacks}.

\section*{Acknowledgments}

This work has been partially supported by the Doctoral College Resilient Embedded Systems, which is run jointly by the TU Wien's Faculty of Informatics and the UAS Technikum Wien. This work was also jointly supported by the NYUAD Center for Interacting Urban Networks
(CITIES), funded by Tamkeen under the NYUAD Research Institute Award CG001 and by the Swiss Re Institute under the Quantum Cities™ initiative, and Center for CyberSecurity (CCS), funded by Tamkeen under the NYUAD Research Institute Award G1104.

\begin{refsize}
\bibliographystyle{ieeetr}
\bibliography{main.bib}
\end{refsize}

\end{document}